\documentclass[runningheads]{llncs}

\usepackage[T1]{fontenc}
\usepackage{amsmath}

\usepackage[tt=false, type1=true]{libertine}
\usepackage[libertine]{newtxmath}
\usepackage[varqu]{zi4}
\usepackage{booktabs}
\usepackage{graphicx}
\usepackage{marvosym}
\usepackage{hyperref}

\begin{document}

\title{Decoupled Single-Mask Annotation Noise Detection via Cross-Sectional Patch Self-Consistency}

\titlerunning{Cross-Sectional Consistency Annotation Noise Detection}

\author{Yinheng Zhu\inst{1}\orcidID{0000-0002-9421-2404} \and
Xiaowei Xu\inst{2}\orcidID{0000-0002-1046-6379}\textsuperscript{(\Letter)}}
\authorrunning{Y. Zhu and X. Xu}
\institute{Tsinghua University, Beijing, China\\
\email{zhuyinheng666@gmail.com} \and
Guangdong Provincial People's Hospital, Guangzhou, China\\
\email{xiao.wei.xu@foxmail.com}}

\maketitle

\begin{abstract}
Vascular computed tomography datasets are commonly annotated only once per scan, yielding the pervasive yet under-addressed problem of \emph{single-mask} annotation noise. Existing solutions either require costly multi-rater fusion or are coupled with network training, preventing explicit auditing of \emph{where} and \emph{why} labels fail.
We introduce a \emph{decoupled} framework for single-mask annotation noise detection that leverages \emph{cross-sectional patch self-consistency} to produce \emph{interpretable} and \emph{auditable} noise evidence. Tubular anatomy exhibits strong cross-sectional recurrence: patches extracted orthogonally along vessel centrelines recur in appearance across locations and subjects. Thus, anatomically similar patches should have consistent masks, and disagreement signals unreliable annotation.
Our method samples cross-sectional patches, retrieves intensity-equivalent neighbours via scalable vector search, and computes a \emph{patch-level noise score} from statistical mask disagreement, yielding explicit image--mask evidence for every flagged region. Aggregating scores produces scan-level \emph{quality maps} for dataset quality assessment or quality-weighted training.
Experiments on the coronary CT dataset validate the detected noise for improving training robustness and reveal systematic annotation biases, i.e., transverse and oblique vessels exhibit 5.1$\times$ higher error rate than axis-aligned structures, with additional correlations to cross-sectional area and intensity. Code is available \href{https://github.com/zhuyinheng/hidden_noise}{here}.

\end{abstract}

\section{Introduction}

Vascular segmentation is a core component in many Computed Tomography (CT) analysis pipelines, yet high-quality vascular annotations are difficult to obtain due to thin \emph{curve-network} anatomy and contrast agent propagation~\cite{zhu2025sparse}. Consequently, real-world datasets often contain localized annotation noise, while being annotated only once per scan because repeated expert labeling is costly. This yields a common but under-addressed setting: \emph{a single noisy mask per image}, which complicates both model training~\cite{zhang_understanding_2017} and dataset quality control~\cite{northcutt_pervasive_nodate}.

Existing noise-handling strategies fall into two categories. (1)~\emph{Multi-rater Fusion} fuses several masks per image~\cite{warfield_simultaneous_2004,liu_istaple_2013,liao_modeling_2022,wu_feda3i_2024,leonardis_quality_2025}, but fundamentally requires multiple raters, making it impractical for vascular CT where datasets almost always contain a single mask per scan.
(2)~\emph{Training-coupled robust learning} incorporates noise handling into network optimization through weighting, disagreement, or structure-aware correction~\cite{liu_adaptive_2022,zhang_cross_2024,shen_co-training_2023,xu_anti-interference_2022,li_superpixel-guided_2021,de_bruijne_study_2021,yao_learning_2023,dong_deep_2024}.
While effective for improving robustness, these approaches are not decoupled from training and cannot provide auditable evidence of label unreliability.
As a result, no existing method supports decoupled and auditable noise localization in the practical single-mask setting.

Our key observation is that tubular anatomy provides natural cross-sectional recurrence across locations and subjects.
We therefore introduce a \emph{cross-sectional patch self-consistency principle}: patches with highly similar appearance should exhibit consistent masks, and strong disagreement indicates unreliable annotation.
To operationalize this idea, we sample orthogonal cross-sectional patches along vessel centrelines, retrieve intensity-equivalent neighbours across scans, and compute a \emph{patch-level noise score} that is directly traceable to explicit patch-pair evidence.
We further aggregate patch scores into a 3D \emph{quality map} for quality assessment (QA) and for practical noise mitigation via quality-weighted training. Our contributions are threefold:
\begin{itemize}
    \item We introduce a {decoupled} framework for annotation noise localization that operates directly on the image--mask pair, enabling practical use in the ubiquitous {single-rater} setting of vascular CT.
    \item We establish a {cross-sectional patch self-consistency principle} for detecting local label inconsistencies, producing inherently {interpretable} and {auditable} evidence via explicit patch-pair comparisons.
    \item We demonstrate that the detected noise exposes {systematic annotation biases} and improves segmentation robustness through {quality-weighted training} on coronary CT angiography.
\end{itemize}

\section{Methodology}

\begin{figure}[tbp]
	\centering
    \includegraphics[width=1.0\textwidth]{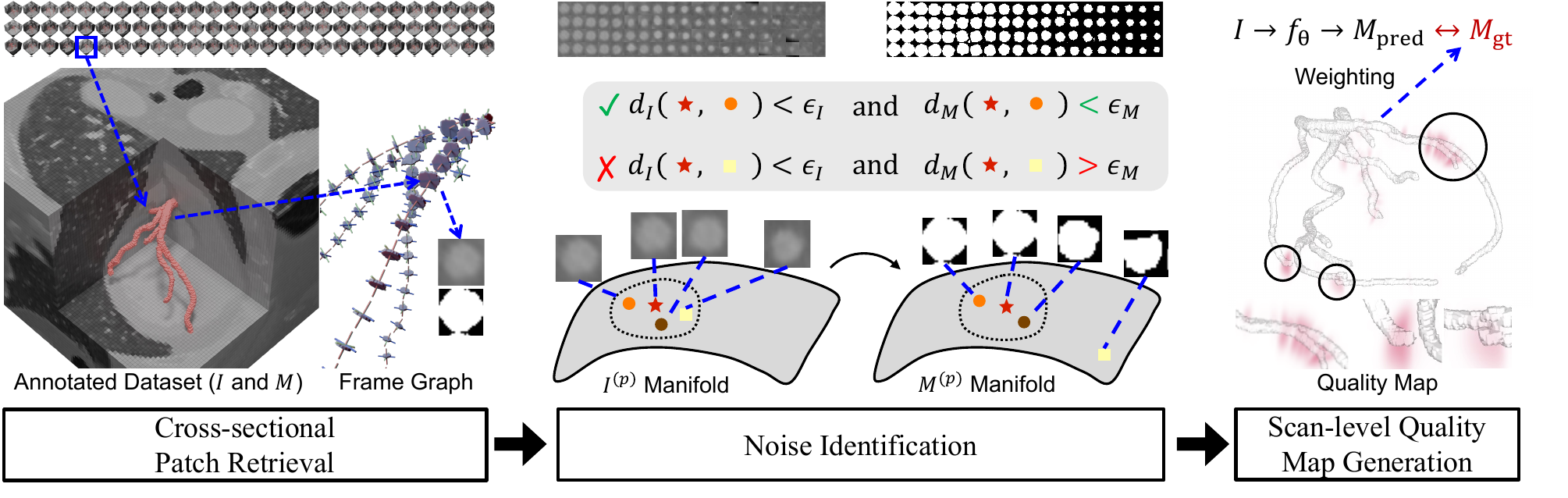}
	\caption{Overview of the proposed method. Left: Cross-sectional patches are extracted along Bishop frames of vessel centrelines, and visually similar patches are retrieved across subjects. Middle: For each patch, its mask is compared with those of its intensity-equivalent neighbours. If any neighbour has a highly similar image but a substantially different mask, at least one annotation is potentially noisy. Right: Pair and patch level scores are aggregated into a scan-level quality map for weighted training.}
	\label{fig:method-overview}
\end{figure}

\subsection{Self-Consistency Principle and Overview}
To make the logical starting point explicit, let $I$ and $M$ denote the CT image and its corresponding segmentation annotation(s). The similarity measures in image and mask spaces are denoted by $d_I$ and $d_M$, respectively.
We first recall the principle underlying classical multi-rater consistency, where segmentations obtained repeatedly from the same image are expected to remain stable~\cite{warfield_simultaneous_2004}:
\begin{equation}
\label{eq:classical}
\text{(Multi-rater Consistency)} \qquad d_I(I_i, I_j)=0 \quad \Rightarrow \quad d_M(M_i, M_j) \leq \epsilon_M.
\end{equation}

However, Eq.~\eqref{eq:classical} is not directly applicable in the single-rater setting, because exact repetition of \emph{whole image} is rare. Considering the key property that vascular structures exhibit strong cross-sectional recurrence, this rationale can be transferred into \emph{cross-sectional patch}. Specifically, when sampled orthogonally along vessel centrelines, cross-sectional image patches~$(I^{(p)}, M^{(p)})$ frequently share highly similar appearances across nearby locations and even across different subjects. Therefore, local image--mask pairs can still be compared meaningfully across the dataset.

Building upon the above, we formulate the cross-sectional patch self-consistency principle as follows. For two patches $(I_i^{(p)}, M_i^{(p)})$ and $(I_j^{(p)}, M_j^{(p)})$,
\begin{equation}
\label{eq:cross}
\text{(Self-Consistency)}
\qquad d_I(I_i^{(p)}, I_j^{(p)}) \leq \epsilon_I
\quad \Rightarrow \quad
 d_M(M_i^{(p)}, M_j^{(p)}) \leq \epsilon_M.
\end{equation}
Accordingly, pairs that satisfy the antecedent but violate the consequent, i.e., high mask disagreement despite near-identical images, are flagged as potentially inconsistent.
The overall pipeline is illustrated in Fig.~\ref{fig:method-overview}, while the following subsections detail each stage.

\subsection{Cross-sectional Patch Retrieval}\label{bishop_frame_graph}

To operationalize Eq.~\eqref{eq:cross} effectively, we need to enlarge the number of near-identical image patch pairs, where the selection of local coordinate frames matters.
We avoid the commonly seen Frenet--Serret frame~\cite{docarmo1976differential} because its torsion term induces unstable rotations, especially in low-curvature segments, reducing the number of near-identical patch pairs and causing numerical instabilities. The Bishop frame~\cite{bishop1975frame} eliminates torsion and yields numerically stable, rotation-free orthonormal directions, making it ideal for consistent patch extraction.

Given a volumetric image $I$ and its segmentation mask $M$, we construct the graph of Bishop frames as follows. We first extract the graph of vessel centreline as described in~\cite{zhu2025sparse}. The Bishop frame of each node is then propagated by parallel transport with zero-twist constraint along the centreline graph, starting from an arbitrary initial frame at the root node.
Once the Bishop frame graph is constructed, we can sample cross-sectional patches, at fixed physical extent to cover the maximum expected vessel diameter, along the normal and binormal plane at each node.
Applying this process to all scans in the annotated dataset $\mathcal{D}$ yields the complete patch set $\mathcal{P} = \{(I_i^{(p)}, M_i^{(p)})\}$.
Algorithmic details of frame construction and propagation are provided in Appendix~\ref{sec:appendix-bishop}.

\subsection{Noise Identification}\label{sec:noise-identification}
\subsubsection{Nearest-neighbour retrieval.}
For each $P_i\in\mathcal P$, we define its neighbourhood as $\mathcal N_i=\{P_\ell\in\mathcal P\mid d_I(I_i^{(p)},I_\ell^{(p)})<\epsilon_I,\ i\neq\ell\}$, computed efficiently via vector search~\cite{douze2025faiss}.
We set $d_I$ as mean squared error (MSE) and $d_M$ as $1-\mathrm{IoU}$ (Intersection over Union). MSE is adopted for scalability: with ${\sim}3{\times}10^6$ patches the retrieval entails ${\sim}10^{12}$ pairwise comparisons, for which structural metrics such as SSIM are computationally prohibitive and incompatible with scalable vector search; the resulting sensitivity to rotation and window-level shifts is analyzed in the failure cases (Fig.~\ref{fig:failure_modes}).
For each neighbour pair $(i,j)$, we compute the distance in image and mask domain, denoted as $d_I(I_i^{(p)},I_j^{(p)})$, $d_M(M_i^{(p)},M_j^{(p)})$ respectively. Aggregating across the dataset, these samples characterize the conditional distribution of mask distance given image distance, i.e., $p(d_M\mid d_I)$.

\subsubsection{Pair-level residual.}
\begin{figure}[tbp]
	\centering
	\includegraphics[width=\textwidth,keepaspectratio]{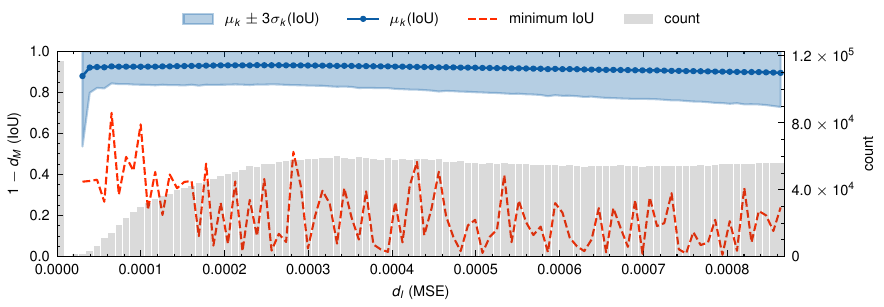}
	\caption{Conditional distribution $p(d_M|d_I)$ and statistics of pairwise residual $r_{ij}$. The blue curve shows the bin-wise mean IoU, the shaded band indicates $\pm3$ standard deviations, the red dashed curve plots the bin-wise minimum IoU, and the gray histogram reports the sample count per bin.}
	\label{fig:residual}
\end{figure}

We summarize $p(d_M\mid d_I)$ by estimating the conditional mean and variance of the mask distance $d_M$ as a function of the image distance $d_I$, and identify pairs whose $d_M$ is unusually high given their $d_I$ (i.e., masks that disagree despite similar images).
Concretely, we partition $[0,\epsilon_I)$ into $K$ equal-width bins $\{B^k\}_{k=1}^K$ according to $d_I$.
For each bin $B^k$, we estimate the conditional baseline of $d_M$ by the sample mean and standard deviation
\begin{equation}
\mu^k \approx \mathbb{E}[d_M\mid d_I\in B^k], \qquad
\sigma^k \approx \sqrt{\mathrm{Var}[d_M\mid d_I\in B^k]}.
\end{equation}
Fig.~\ref{fig:residual} illustrates this calibration.
Under our definition $d_M=1-\mathrm{IoU}$, unusually \emph{large} $d_M$ corresponds to the \emph{lower tail} of $\mathrm{IoU}$ at the same $d_I$ level.
Given a pair $(i,j)$ with $d_I(I_i^{(p)},I_j^{(p)})\in B^k$, we define a similarity-conditioned residual
\begin{equation}
r_{ij}=\frac{\mu^k - d_M(M_i^{(p)},M_j^{(p)})}{\sigma^k + \varepsilon},
\end{equation}
where $\varepsilon$ is a small constant for numerical stability.
Equivalently, $r_{ij}$ is a signed $z$-score analogous to $t$-statistic: a strongly negative value indicates that the observed mask disagreement is anomalously large given the image similarity, i.e., a statistically significant violation of Eq.~\eqref{eq:cross}.
Bins with insufficient samples to estimate $(\mu^k,\sigma^k)$ reliably are excluded.

\subsubsection{Patch-level score.}
Since $r_{ij}$ is a pair-level measure, it does not directly indicate which patch is noisy. We therefore aggregate via the median, asking whether the \emph{majority} of neighbours signal anomalous disagreement:
\begin{equation}
	R_i=\mathrm{median}(\{r_{ij} \mid j\in\mathcal{N}_i\}).
\end{equation}
Under $H_0$ (patch $i$ is clean), residuals scatter symmetrically around zero. If patch $i$ is noisy, its mask consistently disagrees with neighbours, driving $R_i \ll 0$.

\subsection{Scan-level Quality Map Generation}\label{sec:quality-map}
With noise scores computed for all patches, we aggregate them to the scan level to produce a spatially coherent quality map for downstream applications.
We first propagate patch scores to all voxels via Voronoi labeling: each voxel $\mathbf{v}$ is assigned the noise score $R_i$ of its nearest centreline node, yielding a per-voxel residual field $R(\mathbf{v})$.
We then convert $R(\mathbf{v})$ to a voxel-wise quality score via the sigmoid function
\begin{equation}
q(\mathbf{v}) = \frac{1}{1 + e^{-R(\mathbf{v})}},
\end{equation}
and define the final quality map as
\begin{equation}
Q(\mathbf{v}) = 1 - \bigl(1 - q(\mathbf{v})\bigr)\,w(\mathbf{v}),
\end{equation}
where $w(\mathbf{v})$ is a distance-to-centreline weight that decays smoothly away from the vessel.
By construction, $Q(\mathbf{v})\in[0,1]$ decreases monotonically with $R(\mathbf{v})$, while the attenuation term $w(\mathbf{v})$ limits the effect to vessel-adjacent regions to avoid down-weighting background voxels during training.
An example quality map is shown in Fig.~\ref{fig:method-overview}(right). Given any segmentation network $f_\theta$ producing per-voxel predictions $\hat{y}(\mathbf{v})$, we incorporate $Q(\mathbf{v})$ by reweighting a standard voxel-wise loss $\ell(\hat{y}(\mathbf{v}),y(\mathbf{v}))$:
\begin{equation}
\mathcal{L}_{\mathrm{qw}}
=\frac{1}{\sum_{\mathbf{v}} Q(\mathbf{v})}\sum_{\mathbf{v}} Q(\mathbf{v})\,\ell(\hat{y}(\mathbf{v}),y(\mathbf{v})),
\end{equation}
so that voxels with lower estimated annotation quality contribute less to optimization.

\section{Experiments and Results}

\subsection{Experiment Setup}
We evaluate the proposed framework on the ImageCAS dataset~\cite{zeng2023imagecas}, which contains 1000 coronary CT angiography scans with expert annotations. Following the proposed pipeline above, we extract approximately $3\times10^6$ cross-sectional patches with size of $24\times24$ at pixel spacing of 0.125~mm, compute patch-level noise scores $\{R_i\}$ with $\epsilon_I=10^{-3}$ (MSE), corresponding to peak signal-to-noise ratio (PSNR) of~$30\,\mathrm{dB}$ (visually imperceptible), using $K=100$ equal-width residual bins (a trade-off between per-bin variance-estimation stability and bin locality), and construct scan-level quality maps $Q(\mathbf{v})$ that reweight the training loss. We adopt nnUNet~\cite{isensee2021nnunet} with default automatic configuration as the baseline segmentation model and compare standard training with quality-weighted training. The dataset is split into training and test sets at an 8:2 ratio. Segmentation performance is evaluated using both volumetric and vessel-specific metrics. In addition to scan-level Dice score (DSC), we report Dice score on curved planar reformation~\cite{kanitsar2002cpr} view (CPR-DSC) to measure cross-sectional agreement along the centreline, Average Surface Distance (ASD) and Hausdorff Distance at 95th percentile (HD-95) to assess boundary accuracy.
\begin{table}[tbp]
    \centering
    \caption{Quantitative segmentation results with and without quality weighting.}
    \label{tab:results}
    \begin{tabular}{@{}lcccc@{}}
    \toprule
     & DSC $\uparrow$ & CPR-DSC $\uparrow$ & ASD (mm) $\downarrow$ & HD-95 (mm) $\downarrow$ \\
    \midrule
    w/o & 0.812 & 0.801 & 1.63 & 10.27 \\
    w/  & \textbf{0.814} & \textbf{0.815} & \textbf{1.57} & \textbf{9.85} \\
    \bottomrule
    \end{tabular}
\end{table}
\begin{figure}[tbp]
    \centering
    \includegraphics[width=\textwidth,keepaspectratio]{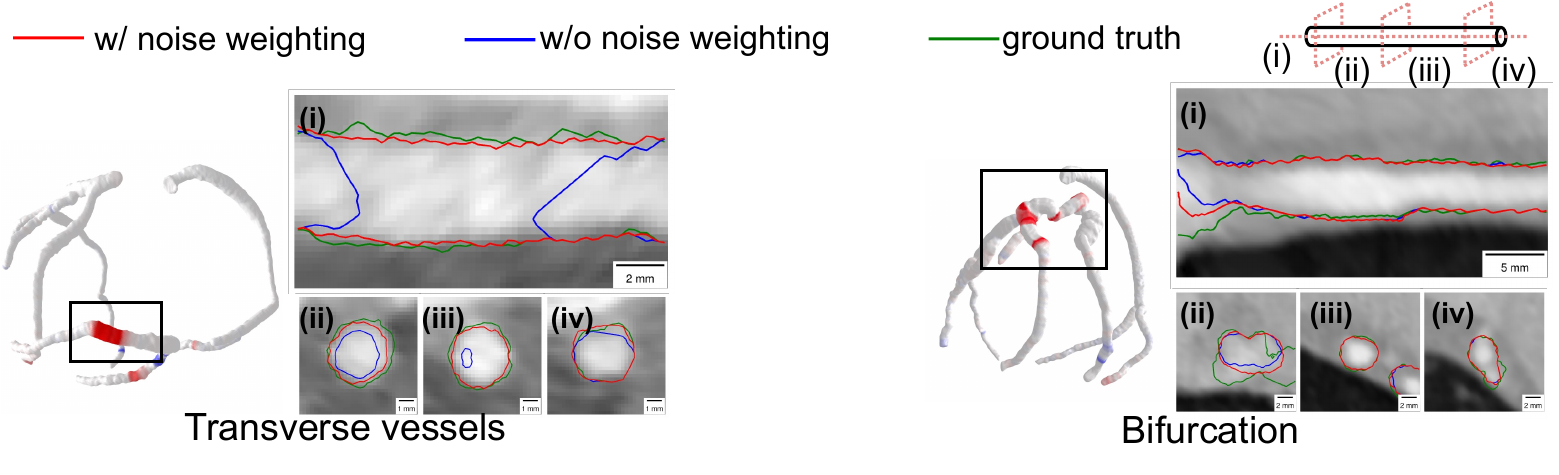}
    \caption{Qualitative comparison on transverse vessels (left) and bifurcations (right), with CPR views (i) and cross-sections (ii--iv).}
    \label{fig:qualitative}
\end{figure}

\subsection{Overall Results}\label{sec:overall-results}
As shown in Table~\ref{tab:results}, the benefits of training with the noise-downweighted dataset are most evident on boundary-sensitive metrics, including CPR-DSC, ASD, and HD-95, indicating improved cross-sectional fidelity and surface regularity. These gains are particularly relevant for vascular segmentation, where geometric accuracy along thin structures and boundaries is often more informative than volumetric overlap alone. Meanwhile, the overall DSC remains comparable, which is expected since DSC primarily reflects foreground volumetric overlap and is less sensitive to localized boundary discrepancies.
Fig.~\ref{fig:qualitative} provides qualitative comparisons in two challenging scenarios. For transverse vessels, whose axes are nearly parallel to the axial imaging plane, quality-weighted training yields contours that align more closely with the reference in both curved planar reformation and cross-sectional views. For bifurcations, where multiple branches converge and boundaries are prone to ambiguity, quality weighting produces smoother and more anatomically plausible delineations, which is particularly relevant for downstream analyses such as stenosis assessment and hemodynamic simulation.

\subsubsection{Computation time.}
The pipeline splits into two stages. Precomputation, which performs top-$k$ nearest-neighbour search over all ${\sim}(3\times10^6)^2$ patch pairs via FAISS~\cite{douze2025faiss} on an NVIDIA RTX 4090, takes approximately 6 hours and is performed once for the entire dataset. After the index is built, noise identification for a single scan of ${\sim}3000$ patches completes in under one minute.

\subsection{Identified Noise and Analysis}
\begin{figure}[tbp]
    \centering
    \includegraphics[width=\textwidth,keepaspectratio]{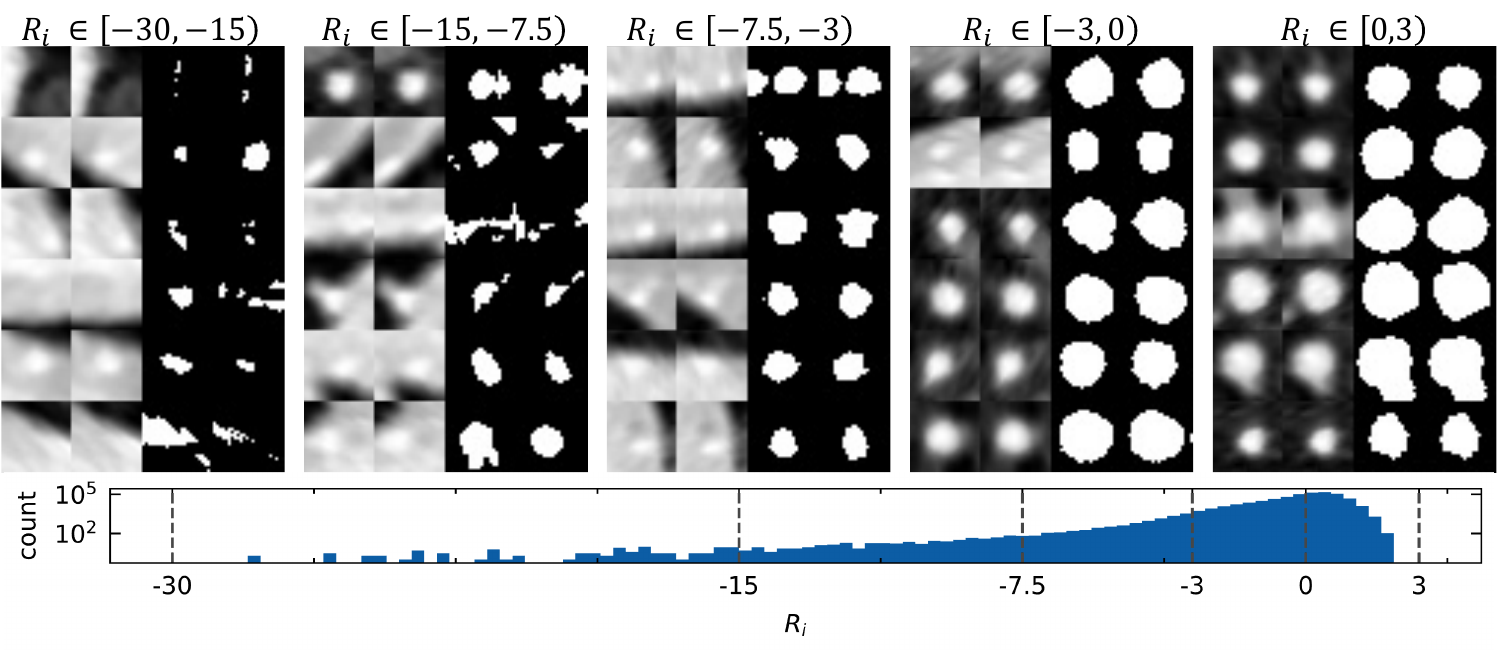}
    \caption{Detected annotation noise across the $R_i$ spectrum. {Bottom}: Histogram of $R_i$ (log-scaled count). {Top}: Representative patch pairs from five $R_i$ intervals.}
    \label{fig:identified_noise_vs_R_i}
\end{figure}

The histogram of $R_i$ (Fig.~\ref{fig:identified_noise_vs_R_i}) exhibits a one-sided fat-tailed distribution, with a small but critical fraction extending to extreme negative values, corresponding to the most inconsistent patches in the dataset.
Representative patch pairs sampled across the $R_i$ spectrum confirm that the score reliably ranks annotation quality: at extreme negative scores ($R_i \in [-30,-15)$), image patches are visually near-identical yet their masks differ drastically, constituting a clear self-consistency violation.
As $R_i$ increases toward $-7.5$, discrepancies become subtler, manifesting as local boundary shifts rather than wholesale segmentation disagreement.
Patches near zero or positive $R_i$ show well-matched masks, confirming that the score correctly identifies clean annotations as well as noisy ones.

\begin{figure}[tbp]
    \centering
    \includegraphics[width=\textwidth,keepaspectratio]{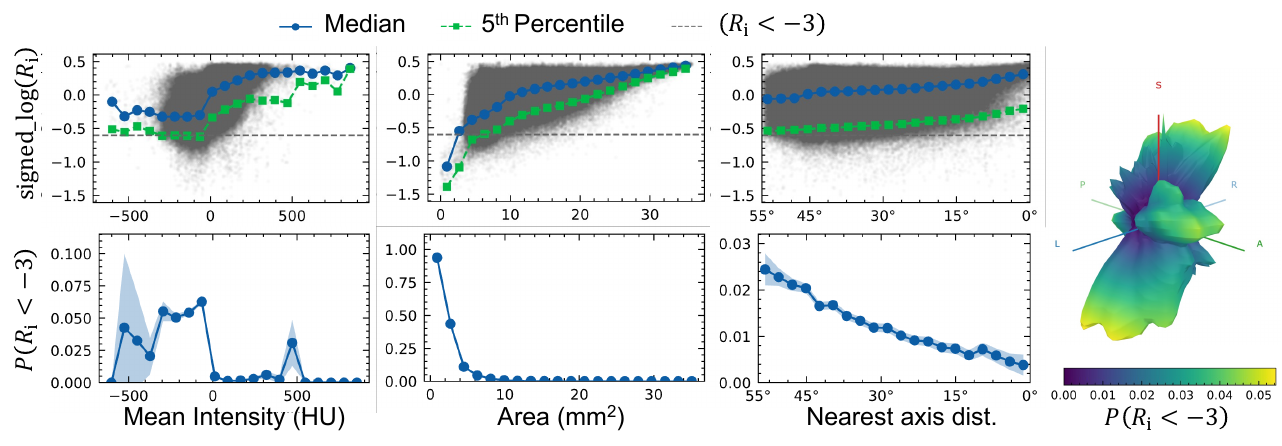}
    \caption{Conditional distribution of $R_i$ versus local attributes.
    The signed-log transform $\mathrm{signed\_log}(R_i)=\mathrm{sign}(R_i)\cdot\log_{10}(1+|R_i|)$ compresses the fat-tailed distribution for visualization. Shaded bands denote 95\% confidence intervals, and wide bands at intensity extremes reflect sparse samples. In the spherical histogram, both radius and color encode $P(R_i < -3)$.}
    \label{fig:R_i_vs_features}
\end{figure}
We further investigate \emph{where} annotation noise concentrates (Fig.~\ref{fig:R_i_vs_features}).
The probability of extreme errors ($R_i < -3.0$) peaks in the low-HU range ($< 0\,\mathrm{HU}$, approximately 5--6\%) and drops in the typical lumen range (50--200 HU), consistent with reduced contrast between vessel wall and background.
Small vessels ($< 2\,\mathrm{mm}^2$) exhibit markedly elevated noise rates, rapidly declining beyond $5\,\mathrm{mm}^2$, likely because thin vessels occupy only a few pixels in axial slices, limiting annotation precision.
Both trends reflect the same underlying condition: reduced boundary discriminability in the annotation interface.
Vessel orientation presents a qualitatively different, directional dependence: we quantify it as the angular distance $\theta$ to the nearest imaging axis (0\textdegree\ for axis-aligned, 55\textdegree\ for maximally oblique).
Spearman correlation yields $\rho = -0.2$ ($p < 0.001$), and the Cochran--Armitage trend test~\cite{agresti2013categorical} confirms a monotonic increase in extreme error rate ($z = -33.31$, $p < 0.001$), rising from $0.44\%$ at $\theta \approx 0$\textdegree\ to $2.24\%$ at $\theta \approx 55$\textdegree\ (a \textbf{5.1-fold} increase).
The spherical histogram demonstrates this pattern: error rates are lowest near each of the six axis-aligned poles and peak in the inter-axis regions, where the vessel axis is far from all three imaging axes simultaneously.
The mechanism is geometric: axis-aligned vessels appear as circular cross-sections in their corresponding imaging plane, whereas oblique vessels appear as elongated ellipses with diffuse boundaries, making consistent delineation inherently harder.

\subsubsection{Failure cases.}
\begin{figure}[tbp]
	\centering
	\includegraphics[width=0.48\textwidth]{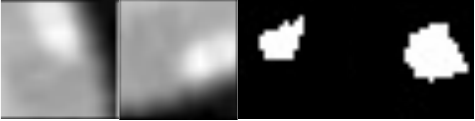}
	\hfill
	\includegraphics[width=0.48\textwidth]{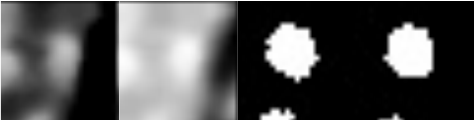}
	\caption{Representative undetected annotation noise. Left: patches near-identical up to rotation. Right: patches near-identical up to window-level shift.}
	\label{fig:failure_modes}
\end{figure}

The dominant failure pattern is false negatives: noise goes undetected when no sufficiently similar counterpart is retrieved. As shown in Fig.~\ref{fig:failure_modes}, this occurs when patches are near-identical only up to rotation or window-level shift, where raw MSE retrieval is sensitive, leaving the inconsistency undetected. Centreline and Bishop-frame errors likewise reduce retrieval \emph{recall}, not \emph{precision}, which our design prioritizes. Recall is recoverable by denser sampling of rotations and shifts at higher computational cost.

\section{Discussion}
\subsubsection{Auditable evidence rather than model behavior.}
The key conceptual innovation is grounding noise detection in \emph{cross-sectional self-consistency} rather than model behavior. Training-coupled single-mask methods infer noise indirectly from optimization dynamics such as loss or feature divergence (e.g., deep self-cleansing~\cite{dong_deep_2024}), leaving it ambiguous whether a flagged region is mislabeled or merely a hard sample, and they consume this signal by excluding or replacing labels inside the training loop. In contrast, our criterion is computed directly from the image--mask pair, so each flagged region traces to a verifiable patch-pair violation and is \emph{auditable}: a suspicious patch is surfaced together with its near-identical neighbours as explicit evidence for human QA, without multi-rater annotation or training interference. The closest network-free analogue is the multi-rater iSTAPLE~\cite{liu_istaple_2013}, which similarly emphasizes a consistency criterion rather than a benchmark comparison.

\subsubsection{Revealing hidden systematic biases.}
Computing the criterion directly from the image--mask pair also enables the discovery of systematic biases otherwise hidden: the dominant orientation dependence follows directly from the slice-by-slice paradigm of tools such as ITK-SNAP~\cite{yushkevich2016itk}, where oblique vessels appear as irregular elongated shapes. Hemodynamic simulation workflows corroborate this by annotating directly on centreline cross-sections~\cite{updegrove2017simvascular}.

\subsubsection{Limitations and future work.}
The method targets tubular anatomy and detects rather than corrects noise, while extension to other tubular structures is straightforward. Our evaluation is confined to a single dataset (ImageCAS), and the cross-sectional recurrence assumption may hold less well on more heterogeneous data; broader cross-dataset validation and explicit label correction are left to future work.

\section{Conclusion}
We present a \emph{decoupled}, \emph{interpretable}, and \emph{auditable} framework for localizing annotation noise from a \emph{single} vascular mask. The method exploits cross-sectional self-consistency: near-identical patches should have near-identical masks, and violations are flagged as noise. Each flagged region is directly traceable to concrete image--mask patch-pair evidence. On ImageCAS, quality-weighted training improves CPR-DSC by 1.4\% and reduces HD-95 by 4.1\%. More importantly, the analysis reveals that \textbf{annotation noise is not random}, and systematic biases exist, correlated with vessel orientation, area, and intensity.

\newpage

\bibliographystyle{splncs04}
\bibliography{reference}

\newpage

\appendix

\section{Bishop Frame Construction and Propagation}
\label{sec:appendix-bishop}

This appendix details the construction of the Bishop frame graph used in Section~\ref{bishop_frame_graph}.

\subsubsection{Centreline extraction.}
Given a binary segmentation mask $M$, we extract the vessel centreline graph as described in~\cite{zhu2025sparse}, via morphological thinning. The result is a set of 3D coordinates $\{\mathbf{x}_i\}$ representing the medial axis of the vascular structure. Connectivity is established by linking nodes that are 26-neighbours in the voxel grid, yielding edge pairs $\{(u_e, v_e)\}$.

\subsubsection{Tangent computation.}
For each node $\mathbf{x}_i$ with an outgoing edge to $\mathbf{x}_j$, the unit tangent is computed as
\[
\mathbf{t}_i = \frac{\mathbf{x}_j - \mathbf{x}_i}{\|\mathbf{x}_j - \mathbf{x}_i\|}.
\]

\subsubsection{Bishop frame propagation via parallel transport.}
Starting from an arbitrary initial frame $(\mathbf{t}_0, \mathbf{n}_0, \mathbf{b}_0)$ at the root node (where $\mathbf{n}_0$ is chosen arbitrarily perpendicular to $\mathbf{t}_0$, and $\mathbf{b}_0 = \mathbf{t}_0 \times \mathbf{n}_0$), we propagate the frame along the centreline by parallel transport. For each subsequent node $i > 0$: (i) Project the previous normal onto the plane orthogonal to the new tangent:
    \(
    \tilde{\mathbf{n}}_i = \mathbf{n}_{i-1} - (\mathbf{n}_{i-1}^\top \mathbf{t}_i)\,\mathbf{t}_i.
    \)
    (ii) Renormalize: $\mathbf{n}_i = \tilde{\mathbf{n}}_i / \|\tilde{\mathbf{n}}_i\|$.
    (iii) Compute the binormal: $\mathbf{b}_i = \mathbf{t}_i \times \mathbf{n}_i$.
This procedure is the discrete rotation-minimizing (parallel-transport) update: it minimizes the frame twist to first order and incurs zero twist in the continuous limit, while avoiding the discontinuities that arise with the Frenet--Serret frame when curvature approaches zero.

\subsubsection{Bifurcation handling.}
At bifurcation nodes where multiple outgoing edges exist, we duplicate the node and propagate separate frames along each branch independently. This ensures that each branch maintains a consistent frame without interference from sibling branches.

\subsubsection{Output.}
The final Bishop frame graph is defined as $G = \bigl(\{\mathbf{x}_i\}, \{(u_e,v_e)\}, \{(\mathbf{t}_i,\mathbf{n}_i,\mathbf{b}_i)\}\bigr)$, providing a stable coordinate system for cross-sectional patch extraction at every centreline location.

\section{Scalable Pipeline Implementation}
\label{sec:appendix-impl}

Processing $N\approx 3\times10^6$ cross-sectional patches requires careful engineering to avoid $O(N^2)$ computational and memory bottlenecks. We describe the key components below.

\subsubsection{Centreline extraction with Numba~\cite{lam2015numba}.}
Vessel centrelines are extracted as described in~\cite{zhu2025sparse}, via morphological thinning of the binary mask. Skeleton parsing (identifying branch points, tracing edges, and building the graph structure) uses Numba JIT-compiled routines operating directly on the flattened voxel array, achieving $>100\times$ speedup over pure Python.

\subsubsection{Cross-sectional patch sampling.}
At each centreline node, a $24\times24$ patch ($3\,\mathrm{mm}$ extent, $0.125\,\mathrm{mm}$ spacing) is sampled on the Bishop-frame plane via PyTorch's \texttt{grid\_sample} with trilinear (image) or nearest-neighbour (mask) interpolation. Batch processing of $\sim 3\times10^3$ frames per scan keeps GPU utilization high.

\subsubsection{Nearest-neighbour retrieval with FAISS~\cite{douze2025faiss}.}
Each patch is flattened into a 576-dimensional vector after per-patch min-max normalization and circular masking. We build a GPU-resident \texttt{IndexFlatL2} index and query for $k=2047$ candidate neighbours per patch in batches of 512 on an NVIDIA RTX 4090; candidates are then thresholded by the image distance $\epsilon_I$ to form the neighbourhood $\mathcal{N}_i$ of Section~\ref{sec:noise-identification}. Index construction and neighbour search constitute the bulk of the one-time precomputation, which takes approximately 6 hours for the full dataset (Section~\ref{sec:overall-results}).

\subsubsection{Out-of-core storage with Zarr~\cite{miles2024zarr}.}
The $N\times k$ index and distance matrices ($\sim50\,\mathrm{GB}$) are stored as chunked Zarr arrays, enabling memory-mapped access without loading full arrays into RAM. Subsequent metric arrays (MSE, IoU, residuals) follow the same scheme.

\subsubsection{Lazy evaluation with Dask~\cite{rocklin2015dask}.}
Bin-wise statistics $\mu^k,\sigma^k$, pairwise residuals $r_{ij}$, and per-patch scores $R_i$ are computed via Dask's lazy array API, which automatically partitions work across CPU cores and streams chunks from Zarr. A 2\,GB in-memory cache reduces redundant I/O.

\end{document}